\documentclass[sigconf, anonymous=false]{acmart}

\usepackage{booktabs} % For formal tables
\usepackage{algorithm,algorithmicx,algpseudocode}
\usepackage{placeins}

\begin{document}
\title{Modeling User Selection in Quality Diversity}

\author{Alexander Hagg}
\orcid{1234-5678-9012}
\affiliation{%
	\institution{Bonn-Rhein-Sieg University of Applied Sciences}
	\streetaddress{Grantham-Allee 20}
	\city{Bonn} 
	\state{Germany} 
	\postcode{53757}
}
\email{alexander.hagg@h-brs.de}

\author{Alexander Asteroth}
\orcid{1234-5678-9012}
\affiliation{%
	\institution{Bonn-Rhein-Sieg University of Applied Sciences}
	\streetaddress{Grantham-Allee 20}
	\city{Bonn} 
	\state{Germany} 
	\postcode{53757}
}
\email{alexander.asteroth@h-brs.de}

\author{Thomas B\"ack}
\orcid{1234-5678-9012}
\affiliation{%
	\institution{Leiden Institute of Advanced Computer Science}
	\streetaddress{Niels Bohrweg 1}
	\city{Leiden} 
	\state{The Netherlands} 
	\postcode{2333 CA}
}
\email{t.h.w.baeck@liacs.leidenuniv.nl}

\renewcommand{\shortauthors}{A. Hagg et al.}

\begin{abstract}
The initial phase in real world engineering optimization and design is a process of discovery in which not all requirements can be made in advance, or are hard to formalize. Quality diversity algorithms, which produce a variety of high performing solutions, provide a unique chance to support engineers and designers in the search for what is possible and high performing. In this work we begin to answer the question how a user can interact with quality diversity and turn it into an interactive innovation aid. By modeling a user's selection it can be determined whether the optimization is drifting away from the user's preferences. The optimization is then constrained by adding a penalty to the objective function. We present an interactive quality diversity algorithm that can take into account the user's selection. The approach is evaluated in a new multimodal optimization benchmark that allows various optimization tasks to be performed. The user selection drift of the approach is compared to a state of the art alternative on both a planning and a neuroevolution control task, thereby showing its limits and possibilities.
\end{abstract}




\copyrightyear{2019}
\acmYear{2019}
\setcopyright{acmcopyright}
\acmConference[GECCO '19]{Genetic and Evolutionary Computation Conference}{July 13--17, 2019}{Prague, Czech Republic}
\acmBooktitle{Genetic and Evolutionary Computation Conference (GECCO '19), July 13--17, 2019, Prague, Czech Republic}
\acmPrice{15.00}
\acmDOI{10.1145/3321707.3321823}
\acmISBN{978-1-4503-6111-8/19/07}

\maketitle

\section{Introduction}

%\todo[inline]{genetic instead of genotypic}

The discovery of new and innovative solutions is a driving force and a guiding principle behind evolutionary optimization. From the design of an unintuitive antenna~\cite{Hornby2006a} and fooling of deep neural networks~\cite{Nguyena} as prime examples of the power of evolutionary search to the development of novelty discovery~\cite{Lehman2011a} algorithms, evolutionary design has been on the forefront of discovering that which is new and of high quality. Bradner~\cite{Bradner2014} showed that oftentimes engineers use optimization algorithms to get a first intuition of what is possible. This reinvigorates the search for new algorithms that optimize for both innovation as well as quality. 

Designers and engineers are not always interested in finding the best but rather have many good options to choose from. Architects generally do not search for the best solution, but rather want to find good solutions that adhere to changing objectives during the design process. Robotics engineers often work in a trial-and-error fashion by continuously readjusting objectives after they discover what is possible and optimal. During this process of discovery, the user wants to be able to influence an optimization algorithm, but rather to have full control. 

Quality diversity (QD) algorithms, which combine optimization and novelty discovery by keeping track of solutions in an archive that is defined by feature based or behavioral diversity measures, are right at that forefront of current research in divergent  optimization. QD searches for diversity in terms of a solution's expressed behavior~\cite{Cully2015} or features describing the expressed genome~\cite{gaier2018data}. This makes QD an excellent search algorithm for design processes. 

In order to use QD as an innovative design procedure, the user needs to be able to interact with it. One way of interacting is described in the ``design by shopping'' paradigm~\cite{Balling1999}, where the user selects from generated designs. Yet the stochastic nature of mutation in an evolutionary algorithm may cause it to drift away from a selection of solutions that is preferred by the user. This \textit{user selection drift} needs to be controlled if QD is to be integrated into a real world design process. In this work the user's selection of preferred solutions is modeled, thereby allowing adjustments to the objective function based on a simple penalty measure. 

QD algorithms perform multimodal optimization (MMO), in the sense that they search for many solutions, or local optima, of a problem. In ~\cite{Preuss2015} the author describes finding all basins of attraction, regions in the search space where a local search algorithm converges, as one of the tasks in MMO. In QD diversity is measured not in genotypic space but rather in behavior space. The idea that the representations used in architecture and engineering have to be abstracted towards a representation that is optimizable~\cite{Bradner2014} seems to be counterintuitive to the engineer, yet a necessity for computer \textit{aided} design. In this work, selection will therefore take place on the basis of behavioral aspects, while optimization is based on genotypic parameters. The feasibility of modeling this selection is evaluated for two types of genotype to phenotype to behavior translations. 

The design and optimization process is one of both understanding requirements and finding optimal solutions to those requirements. Computer aided ideation has the potential to reduce the number of iterations in a design process by automated generation of design candidates~\cite{Bradner2014}. In principle, QD was shown to be usable in the design by shopping paradigm~\cite{Hagg2018}. 

In this paper the question is answered whether a user's selection can be modeled to properly constrain QD according to that selection. This is a necessary step to turn QD into an innovation aid in a human-computer interacting optimization loop, assisting the designer in decomposing both the problem of understanding requirements while simultaneously finding solutions. 

The remainder of this paper is organized as follows: after a short introduction into QD, current challenges are highlighted, how to integrate QD into real world design processes through user selection. A model that captures a user's decision to select preferred solutions is then introduced and formalized, and evaluated in a new multimodal benchmark that allows performing various tasks in the same domain. The multimodal benchmark consists of a planning and a control task, highlighting the differences in genotype to phenotype and phenotype to behavior mapping. By using different representations in the tasks, the quality of capturing the user decision is compared to that of a related method, showing that the robustness of this new approach is improved. 

\section{Background}
\begin{figure*}[t]
	\centering
	\includegraphics{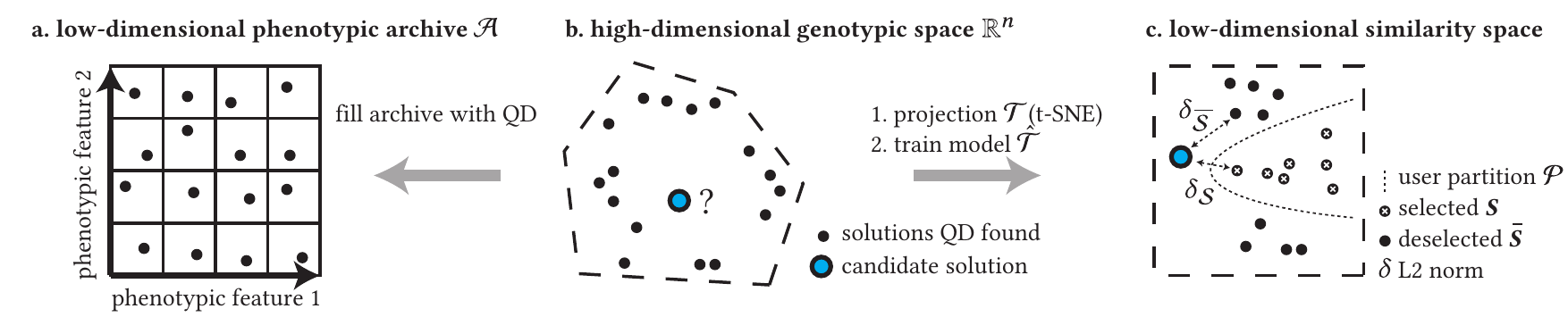}
	\caption{QD searches through genotypic space $\mathbb{R}^n$ (b) to fill an archive $\mathcal{A}$  of diverse, high-performing phenotypes (a) in a low-dimensional phenotypic (or behavior) space. The genotypic dimensionality $n$ can be very high. By projecting the archive's members onto a low-dimensional similarity space (c), the user's selection can be modeled. The projection model $\hat{\mathcal{T}}$ allows making comparisons of candidate solutions to the user selection $\mathcal{S}$ on the hypersurface by using an L2 norm.}
	\label{fig:hyper}
\end{figure*}

QD distinguishes itself from other optimization algorithms by maintaining a diverse archive of high performing solutions, measuring diversity in terms of behavior in what is called a behavior or feature space. Originally the idea behind Novelty Search \cite{Lehman2011a} is to maintain a permanent archive of individuals that shows behavior that is novel at the time of discovery. Adding local competition allowed the algorithm to perform in settings where the novelty measure is used on the genotype, as there is no guarantee a solution is functional. Novelty Search with Local Competition (NSLC)~\cite{Lehman2011} firstly combined quality and diversity, and a few years later Multidimensional Archive of Phenotypic Elites (MAP-Elites)~\cite{Cully2015} was introduced. The main differences between the methods are the way they define an archive and how they select the next population of solutions. In NSLC the archive is unstructured and consists of a historical trace of best, diverse individuals. The most novel and locally high performing individuals in the population, which is kept separate from the archive itself, are selected for the next generation. Niches in the archive are slowly built up over the duration of the algorithm. In MAP-Elites the archive is a predefined and fixed discretization of a user defined descriptor space, represented as a lattice, where every lattice point is an $n$-dimensional object (Fig.~\ref{fig:hyper}b). The next generation is randomly selected from the archive with the intend to fill it over time.

QD algorithms in general follow the pattern described in \cite{Pugh2016,Cully2017} and reiterated in Alg.~\ref{alg1}. After initializing a population of solutions and setting an evaluation budget, parents are selected based on a scoring scheme, which could be a novelty, competition or curiosity measure or a combination thereof, and offspring is created. After initializing the population, the performance of the offspring is evaluated and their descriptors calculated. Depending on the type of archiving, the individuals can be added to existing cells in the archive, or new cells can be created. Finally, the scores are updated.
\begin{algorithm}
	\caption{Quality diversity (QD)}
	\label{alg1}
	\begin{algorithmic}
		\State \textbf{Initialize} population
		\For {\textcolor{black}{{iter $ = 1 \to \text{generations budget} $}}}
		\State \textcolor{black}{\textbf{Select} parents to form offspring based on scoring scheme}
		\State \textcolor{black}{\textbf{Evaluate} performance and descriptor of offspring}
		\State \textcolor{black}{\textbf{Add} individuals (potentially) to archive $\mathcal{A}$}%, according to existing solutions}
		\State \textcolor{black}{\textbf{Update} novelty, competition or curiosity scores}
		\EndFor
	\end{algorithmic}
\end{algorithm}
QD was applied in a Bayesian optimization context for the first time in \cite{gaier2018data}. Surrogate-assisted illumination (SAIL), based on the MAP-Elites algorithm by \cite{Cully2015}, finds many optimal designs while using only a small number of real objective evaluations, increasing the efficiency of MAP-Elites by three orders of magnitude. Diversity is defined in terms of shape features rather than behaviors, and SAIL was applied on shape design optimization.

The integration of QD into real world design processes is a recent development, although evolutionary algorithms have been developed with this integration in mind. In \cite{Parmee2000}, an interactive strategy is developed for innovative conceptual design in real world design processes. The authors identify requirements for such a design process, such as the ability to efficiently sample the design space, the ability to change constraints and objectives, identification of high performance regions, and capturing specific design knowledge through interaction with a designer. The authors introduce cluster oriented genetic algorithms to identify high performance regions, and decompose the optimization problem. 

In~\cite{Hagg2018} the initial intuition of the design space produced by QD was analyzed by clustering QD's resulting genomes in an unsupervised manner, forming design classes and prototypical representatives. The discovery that the behavior based niches in the QD archive contain a relatively small set of design classes is in line with the analysis in~\cite{Vassiliades2018}. That work showed that the elite hypervolume, the part of the genotype space that contains the archive's elites, is often less spread out than the elites are in behavior space. The concise representation in~\cite{Hagg2018} lets a user select interesting classes without being overwhelmed by the large number of solutions produced by QD. To influence the continuation of the QD algorithm, the authors reseeded its archive with the selected results, similar to~\cite{Woolley2014}, thereby forcing QD to start searching around the selection. This way, QD can be used in a design by shopping loop~\cite{Balling1999}.

The approach takes into account the problems with measuring distances in high-dimensional space \cite{beyer1999nearest}. Solutions can be clustered into classes around local optima by introducing a similarity space, calculated by projecting the solutions in the archive with t-distributed stochastic neighborhood embedding (t-SNE)~\cite{VanDerMaaten2008}, which retains global and local structure well for high-dimensional objects. Although QD now starts its search around selected solutions, the unconstrained objective function still allows QD to find solutions within non-selected regions. As a design process consists of many design decisions, it is unclear how the seeding approach would be successful in that case. The subspace that contains solutions fulfilling all of the user's decisions becomes smaller and more complex as more design decisions are added, but QD is divergent and will discover solutions outside of the selection. QD has to be constrained by modeling the user's selection decisions. 

The genotype based selection process from~\cite{Hagg2018} is subject to another problem. When a non-linear genotype to phenotype mapping is used or the phenotype's behavior is reactive, e.g. when evolving neural network controllers or producing shapes from neural representations~\cite{Clune2004}, small changes in the genotype can lead to large changes in the phenotype/behavior. For a designer this would be unexpected, as they assume that once solutions are selected, the continuation of the design process will produce similar solutions. Because similarity is measured in genotype space, such a non-linear mapping causes unexpected behaviors to be discovered, which would be counterintuitive to the designer.

In the following section a model of the user's selection is introduced, which allows comparing candidate solutions to the selected ones.

\section{User Decision Hypersurface Model}
A user's selection needs to be modeled to properly and continuously constrain the optimization process. QD fills its niching archive based on the behavior or phenotype of solutions (Fig. \ref{fig:hyper}a), but the search itself still takes place on the level of the genotype (Fig. \ref{fig:hyper}b). All search constraints need to be defined on the genotype. However, the genotype often has many parameters, causing distance metrics like Euclidean distance to become meaningless, which can happen for as few as 10-15 dimensions~\cite{beyer1999nearest}. \cite{Shaham2017,Hagg2018} showed that when the dimensionality of genotypic space is reduced with t-SNE~\cite{VanDerMaaten2008}, more structure in the objective landscape can be discovered by clustering techniques that are based on the notion of distance. This provides evidence that distances measured in the similarity space resulting from t-SNE are more effective in describing genotypic similarity than those measured in the original genotypic space.

The user's decision is modeled in the similarity space, which is created based on a t-SNE mapping applied to solutions that have been placed in the QD archive at the time the user makes their decision. The archive members form a lattice, which can be seen as a representative of a hypersurface that unfolds in $\mathbb{R}^n$ as new solutions are added to empty cells and moves through $\mathbb{R}^n$ as solutions in cells are exchanged. Every cell in the archive can exclusively contain points of a subspace of $\mathbb{R}^n$, but a particular instance of an archive can be mapped into a lower dimensional space that retains the structure of the archive, while allowing to measure distances on this approximation of the decision hypersurface.

The user decision hypersurface model (UDHM) is formalized as follows:

\begin{align*} 
\mathcal{A} = \{ \mathbf{x}_1,\dots, \mathbf{x}_m \}, \mathbf{x} \in \mathbb{R}^n &: \text{QD archive with $m$  points}\\
\mathcal{H} = \text{span}(\mathcal{A}) &: \text{decision hypersurface}\\ 
\mathcal{T}: \mathcal{A} \rightarrow \mathcal{A}' \subseteq \mathbb{R}^d &: \text{projection into similarity space}\\
\hat{\mathcal{T}}: \mathcal{H} \subseteq \mathbb{R}^n \rightarrow \mathcal{H}' \subseteq \mathbb{R}^d &: \text{projection model}\\
\delta:\mathbb{R}^d \rightarrow \mathbb{R}  &: \text{distance measure, e.g. L2 norm}\\
\mathcal{S}, \overline{\mathcal{S}} &: \text{selected/deselected solutions.}\\
\mathcal{P} = \{\mathcal{S},\overline{\mathcal{S}}\}&: \text{binary selection partition.} \\
\mathcal{M} = ( \hat{\mathcal{T}}, \delta, \mathcal{P} ) &: \text{user decision hypersurface model}
\end{align*}

The span of the points in the archive $\mathcal{A}$ defines the decision hypersurface $\mathcal{H}$ that is projected into similarity space $\mathbb{R}^d$ using t-SNE $\mathcal{T}$.  A dimensionality of $d=2$ is chosen for visualization purposes and because the projection method, t-SNE, has been robustly tested for this case. 

t-SNE itself does not provide a model, but merely maps high-dimensional points onto a lower-dimensional space. It is sensitive to local optima of its cost function, and therefore is not guaranteed to produce the same result in multiple runs~\cite{VanDerMaaten2008}. Due to this sensitivity and the performance cost of the calculation, t-SNE's mapping is modeled using $d$ separate Gaussian Process (GP) regression models~\cite{rasmussen2004} for each of the $d$ similarity space coordinates, resulting in the projection model $\hat{\mathcal{T}}$. The GP models are trained using the archive members based on which a decision is made, and thus fully describes the knowledge that was present at that point in time. The models' isotropic Mat\'ern kernel is applied on the coordinates of the model's training samples in $\mathbb{R}^d$. The length scale priors to training are set to the mean Euclidean distance between the samples in $\mathbb{R}^d$. The projection model $\hat{\mathcal{T}}$, consisting of the two GP models, prevents any unexpected drift that can be caused by changes that would arise due to recalculation of the t-SNE mapping. 

The user decides which solutions they would like to further investigate, represented by the binary selection partition $\mathcal{P}$ (Fig.~\ref{fig:hyper}c). The UDHM $\mathcal{M}$, which contains the projection model $\hat{\mathcal{T}}$, a metric $\delta$ and the user's selection $\mathcal{P}$ allows us to compare candidate solutions to the user's selection based on their genotypes, in similarity space. A decision metric can now be defined that determines whether a candidate solution is closer to $\mathcal{S}$ or to $\overline{\mathcal{S}}$.

\subsection{User Selection Drift}
\label{sec:decisionmetric}

\begin{figure}[htbp]
	\centering
	\includegraphics{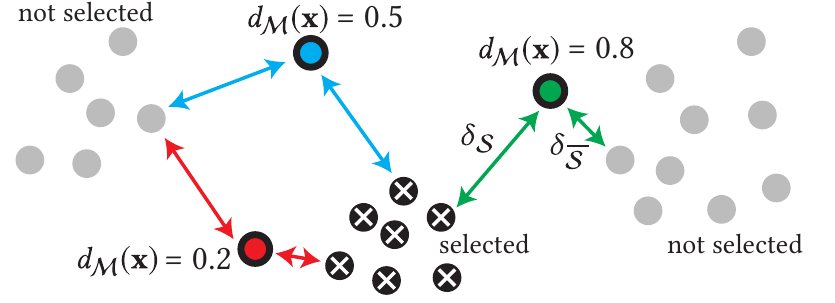}
	\caption{User selection drift $d_{\mathcal{M}}$ is based on the distance to the closest selected point $\delta_{\mathcal{S}}$ and the distance to the closest deselected point $\delta_{\overline{\mathcal{S}}}$.}
	\label{fig:decisionmetric}
\end{figure}

In order to measure how close a solution is to the user selection, a metric called \textit{user selection drift} $d_{\mathcal{M}}$ is introduced. It determines the distance between a candidate solution $\textbf{x}_c$ and the selection $\mathcal{S}$:
\begin{align*} 
\delta_{\mathcal{S}} =  min(\delta(\mathcal{T}(\textbf{x}_c), \mathcal{S})) &: \text{min. distance to selected} \\
\delta_{\overline{\mathcal{S}}} =  min(\delta(\mathcal{T}(\textbf{x}_c), \overline{\mathcal{S}})) &: \text{min. distance to deselected} \\
d_{\mathcal{M}}(\textbf{x}_c) = {\delta_{\mathcal{S}} \over (\delta_{\mathcal{S}} + \delta_{\overline{\mathcal{S}}})}, 0 \le d \le 1 &: \text{normalized user selection drift}
\end{align*}
$d_{\mathcal{M}}$ measures the distance between $\textbf{x}_c$ and the closest point in $\mathcal{S}$, $\delta_{\mathcal{S}}$, and between it and the closest point in $\overline{\mathcal{S}}$, $\delta_{\overline{\mathcal{S}}}$ (Fig.~\ref{fig:decisionmetric}). The normalized user selection drift equals 0 when it exactly matches a selected point and equals 1 when the candidate has the same coordinates as a deselected point. The UDHM and user selection drift can be used to augment the objective function with, for example, a penalty metric.

%\FloatBarrier
\subsection{User Driven Quality Diversity}
\label{sec:udqd}

To make use of the UDHM, QD is extended by including the UDHM to the user-seeded version of MAP-Elites~\cite{Hagg2018}. This user driven quality diversity (UDQD) algorithm is interactive, although it is evaluated based on predefined rules that represent the user's decision.

\begin{algorithm}
	\caption{User driven quality diversity (UDQD)}
	\label{alg3}
	\begin{algorithmic}
		\State \textcolor{gray}{\textbf{Initialize} population}
		\For {\textcolor{black}{\textit{$\text{design iter} = 1 \to \text{design iterations budget}$}}}
		\For {\textcolor{gray}{{iter $ = 1 \to \text{generations budget} $}}}
		\State \textcolor{gray}{\textbf{Select} parents to form offspring based on scoring scheme}		
		\State \textcolor{gray}{\textbf{Evaluate} performance and descriptor of offspring}
		\State \textcolor{gray}{\textbf{Add} individuals (potentially) to archive $\mathcal{A}$}%, according to existing solutions}
		\State \textcolor{gray}{\textbf{Update} novelty, competition or curiosity scores}
		\EndFor
		\State \textcolor{black}{\textbf{Project} archive $\mathcal{A}$ onto $\mathcal{A}'$ with t-SNE.}
		\State \textcolor{black}{\textbf{Train} projection model $\hat{\mathcal{T}}$ with training pairs $(\mathcal{A},\mathcal{A}')$}
		\State \textcolor{black}{\textbf{Determine} partition $\mathcal{P}$ \textbf{by the user} with selection represented by $\mathcal{S}$}
		\State \textcolor{black}{\textbf{Adjust} objective function with penalty $p$} \Comment{Use UDHM} 
		\State \textcolor{black}{\textbf{Assign} $\mathcal{A} \leftarrow \mathcal{S}$} \Comment{Seed QD archive}
		\EndFor
	\end{algorithmic}
\end{algorithm}

UDQD is formalized in Algorithm~\ref{alg3}, with the adjustments to Algorithm~\ref{alg1} in black. A design iterations budget is introduced, although this could be an open loop as well. After an archive is created in the inner QD loop, the similarity coordinates are determined using t-SNE, and the projection model $\hat{\mathcal{T}}$ can be trained using the pairs from $(\mathcal{A},\mathcal{A}')$. The user can now determine which individuals are of interest ($\mathcal{S}$) and which are not ($\overline{\mathcal{S}}$). This decision is then turned into a penalty function and the objective function is adjusted accordingly:
\begin{align*} 
d_{\mathcal{M}}(\textbf{x}), w_p &: \text{drift penalty, penalty weight} \\
f'(\textbf{x}) = f(\textbf{x}) \cdot (1 - w_p \cdot d_{\mathcal{M}}(\textbf{x})) &: \text{adjusted objective (maximization)}
\end{align*}
The penalty, a value between 0 and 1, is used to penalize a solution's fitness. The UDHM determines how far a solution might be mutated away from a known selected solution by measuring the distance to solutions from the state of the archive at the time the decision was made. The measurement is taken in similarity space, in our case created by t-SNE. As t-SNE compresses the parameters into a lower dimensional space, the space is not homogeneous and measurements in that space cannot be translated to a Euclidean measurement in parameter space. Neither a simple threshold can be used to determine whether a candidate is closer to one solution or the other, nor can it be assumed that there is a universal penalty function that works in all domains and in all t-SNE projections. Therefore, a simple linear penalty function is used, which is 0 at points in $\mathcal{S}$ and equals weight $w_p$ at a known point in $\overline{\mathcal{S}}$, to influence the objective function. The penalty is used to scale the objective function. It is dependent on the range of the fitness function values as well as the structure of the hyperspace. For this reason the penalty has to be parameterized for each domain and task. A solution's fitness is penalized when it is unlikely to belong to $\mathcal{S}$, but it will still be accepted when there is no alternative solution, because in the conceptual phase of an engineering task it is better to show alternatives in a niche than showing no solutions at all. By showing the user selection drift per niche, the user can be informed about its supposed distance to the selection.

In the following Section, the model is evaluated for a linear genotype to phenotype mapping as well as a nonlinear, reactive case. Both representations' behaviors are measured in a similar behavior space.

\section{Evaluation}

The main hypothesis is that QD using an objective function that is adjusted with a penalty based on UDHM shows less user selection drift than when seeding it with selected solutions. Evaluation takes place in two tasks that are both defined in the same domain using the same selection criteria, objective function and diversity measure. Because design decisions are based on the way a solution is expressed in the problem domain, which can be a shape or a behavior, the selection process that is introduced is based on the behavior of solutions. Evaluation takes place for two tasks that can be more easily quantified than a design optimization task, but is an analogy to compare directly and indirectly encoded representations.

\subsection{Experimental Setup}

\begin{figure}[htbp]
	\includegraphics{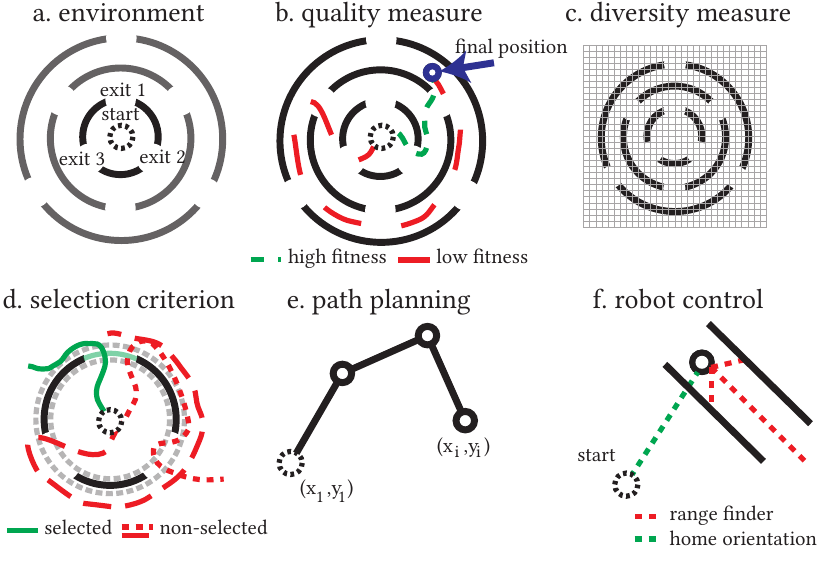}
	\caption{The multimodal maze (a) has a starting location in the center and multiple exits. Due to the symmetry of the problem, no mode, a path that exits the maze is preferred. The quality measure is defined by the \textit{shortest path to the end position of the path taken} (b), so escaping the maze is not explicitly modeled. Solution diversity is induced by aligning the archive with the maze itself (c). The user selects the paths that exit the inner ring (d). When a path reenters the inner ring and exits through another gate, the path is deselected. The two tasks are shown in (e) and (f).}
	\label{fig:maze}
\end{figure}

The multimodal maze presented here is an alternative to the QD gauntlet introduced in \cite{Pugh2016a}. The maze contains three rings, each containing three exits (Fig.~\ref{fig:maze}a). In this domain, solutions are expressed as paths through the maze. The maze is symmetric and has multiple possible, similarly shaped paths through all exits. These paths represent multiple basins of attraction.

Initially, a systems designer might not be able to fully describe the objective or constraints of the controllers. This is simulated by using a more generalized objective function to find a wide variety of efficient solutions. The objective function, or quality measure, that is to be minimized is defined as the length of the path to the final position in a path and does not explicitly model the objective to escape the maze (Fig.~\ref{fig:maze}b). This represents the fact that solutions are designed within the space of optimal solutions, but the final objective is not known or formalizable in advance. A grid of cells in the maze is used as a diversity  measure (Fig.~\ref{fig:maze}c). The diversity archive is aligned to the environment for simplicity. After filling the archive with an initial set of solutions, the user will select the solutions that escape the inner ring of the maze through a particular exit (Fig.~\ref{fig:maze}d).

\noindent The first task is a path planning problem in which the genotype to phenotype mapping is very simple. A path planning solution is encoded by a sequence of seven (x,y) nodes that, when connected, form a path (Fig.~\ref{fig:maze}e). The range of the node coordinates is limited between -200 and 200. Diversity is aligned to the maze and the encoding ensures that solutions showing the same phenotypic behavior, e.g. take similar paths, are also similar in terms of their genotypes. For this task phenotypes and behavior are the same. Because a small change in the genotype causes a small change in the phenotype, behavioral similarity matches genotypic similarity and the UDHM should perform well. 

In the second task, optimization of neural robot controllers, small changes can lead to large changes in the phenotype due to the non-linearities in and reactivity of the neural controller. Solutions are evaluated in the same maze as the path planning task, using the simulation that was created in~\cite{Mouret2011}. A robot is equipped with three range finders that are able to detect the distance to the nearest walls, and a home beacon that detects the quadrant in which the direction to the start position of the robot lies (Fig.~\ref{fig:maze}f). It is controlled by a derivation of a recurrent Elman~\cite{ElmanJ.1990} network which controls two outputs: forward/backward and rotational movement. The network contains five hidden neurons and five context neurons, whereby the weights to the context layer are evolved as well, for a total of 92 weights. The weights' range lies between -3 and 3. The simulation is run for 1000 time steps.

The paths taken by the robots are not directly encoded in the genome, but result from the interaction of the phenotype, the neural network, with the environment. Therefore, similar controllers could display different behavior, which should make the comparison of solutions by their genotypes less effective. The task is therefore useful to show limits of the comparison in similarity space when using a non-linearly coupled genotype, phenotype and behavior. With this final task the limits of the approach are tested. The following introduces the maze domain, and then separately describes the two tasks and the configuration of the UDQD algorithm.

The MAP-Elites archive is set to contain 900 (30 x 30) elites to ensure that the original t-SNE implementation is able to converge within a relatively short time. For larger archives it is recommended to use Barnes-Hut t-SNE~\cite{VanderMaaten2014} which is able to deal with much larger data sets. MAP-Elites is initialized with 2000 (planner) or 200 (controller) solutions. Each configuration is repeated six times for all three exits to account for stochastic effects. The initial orientation of the robot is changed by 60\textdegree~steps starting at 30\textdegree~between runs of the algorithm. In the path planning task the population is initialized using a normal distribution with a small $\sigma$ to ensure that most initial paths are within the center area of the maze to prevent many invalid solutions. In the control task the controller weights are chosen from a space filling Sobol sequence~\cite{niederreiter1988low}. Each initial archive created with a MAP-Elites run takes 8192 generations in the path planning task, and 2048 in the control task. In every generation, 32 children are created through normally distributed mutation with $\sigma = 5\%$ for the planning and $\sigma = 1\%$ for the control task. Parent selection is done by randomly choosing solutions from the archive. 

The hyperspace GP models for both coordinates use an isotropic Mat\'ern kernel. The length scale hyperparameter prior \cite{rasmussen2004} is set to the mean Euclidean distance between the points in the archive in $\mathbb{R}^n$. The length scale prior can make or break the model's ability to correctly compare solutions, as the training of the GP models will not converge. When the length scale is too short, the penalty will be too high for candidate solutions close to $\mathcal{S}$. When the length scale is too long, the penalty will be too low for those close to $\overline{\mathcal{S}}$. After training and constituting the UDHM model, MAP-Elites is run for 4096 generations in the second constrained iteration based on the user selection. 

\subsection{Selection on Hypersurface}

\begin{figure}[htbp]
	\centering
	\includegraphics{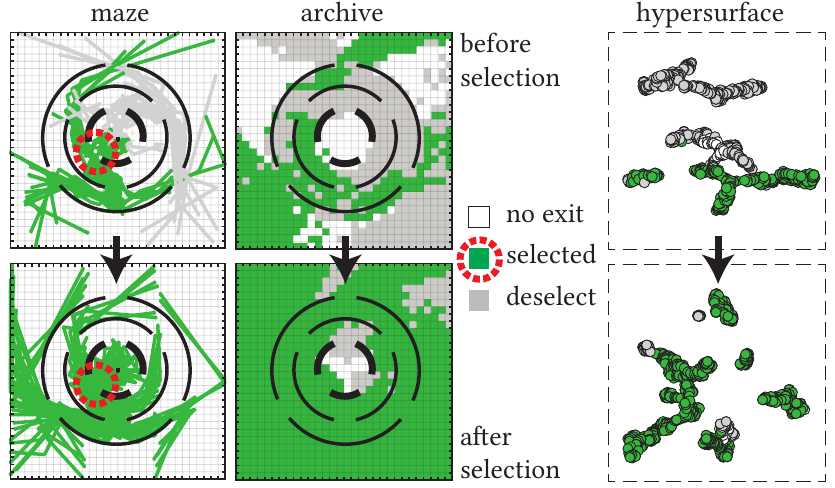}
	\caption{Path planning task before (top) and after (bottom) selection of the 3rd exit. Left: example paths (every 5th). Center: QD archive with color assigned depending on whether exit was selected. Right: points projected onto the decision hypersurface.}
	\label{fig:mapelites}
\end{figure}

Here are some examples from the results from both tasks in more detail. User selection is based on which exit in the inner ring is used by a solution. Fig.~\ref{fig:mapelites} shows example paths that are found by QD in the first iteration without user selection in the path planning task (top row). Solutions that do not get out are marked in grey. The user selects the solutions that take the preferred exit in the \textbf{inner} ring of the maze, in this case the lower left exit, marked in red/green. The hypersurface, or similarity space~\cite{Hagg2018}, shows that solutions that are close together tend to take the same exit.

An example archive that is produced by UDQD after selection of the lower left exit and continuing MAP-Elites for another 4096 generations is shown in the bottom row of Fig.~\ref{fig:mapelites}. MAP-Elites uses the adjusted objective function (Section \ref{sec:udqd}), resulting in an archive that is mostly filled by paths that take the selected exit. The most right column of Fig.~\ref{fig:mapelites} shows the archive's contents projected onto the hypersurface before and after MAP-Elites has been adjusted for selection. Solutions in the archive are well separated according to their behavior, the exit they took in the inner ring, which is to be expected with a direct genotype to phenotype mapping.
\begin{figure}[htbp]
	\centering
	\includegraphics{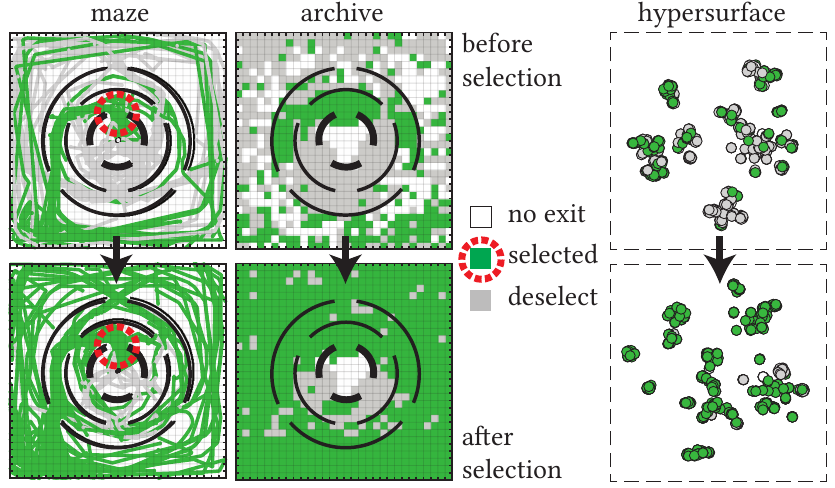}
	\caption{Neurocontrol task before (top) and after (bottom) selection of the 1st exit. Left: example paths (every 5th). Center: QD archive with color assigned depending on whether exit was selected. Right: points projected onto the decision hypersurface.}
	\label{fig:mapelites2}
\end{figure}

The solutions in the robot neurocontrol task are not as well separated on the hypersurface (right column of Fig.~\ref{fig:mapelites2}). The paths the robots take look qualitatively different from the ones in the path planning task and the archive does not fill up in the same way. Yet the user selection is still effective, as can be seen by the archive being filled up by almost entirely by solutions that take the preferred exit.

\subsection{Influence of Penalty Weight on Drift}

\begin{figure}[h!tbp]
	\centering
	\includegraphics{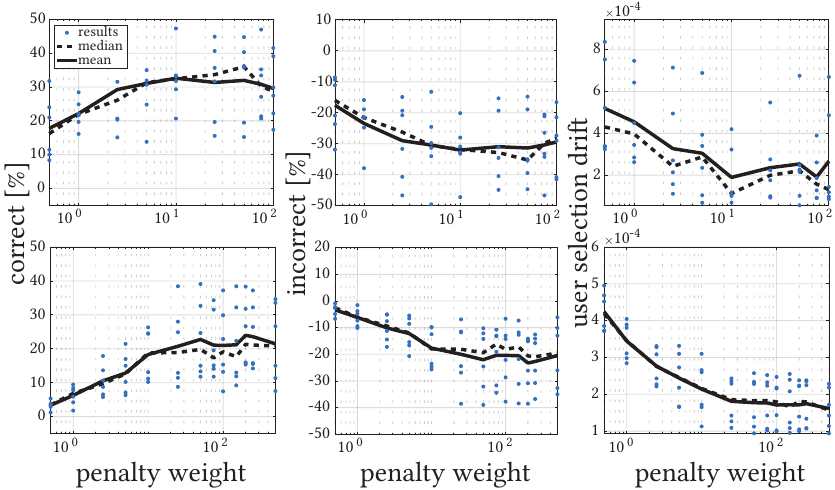}
	\caption{Influence of penalty weight derived from UDHM drift on percentage of selected and deselected exits for the path planning (top) and neurocontrol task (bottom). Left: percentage point increase of correct paths using UDHM compared to running QD without selection. Center: percentage point decrease of number of paths taking wrong exit (compared to QD without selection). Right: decrease in user selection drift when penalty weight is increased.}	
	\label{fig:parameterization}
\end{figure}

The penalty weight's efficacy is evaluated using UDQD with a mutation distance of 5\% of the range of the genes for 4096 generations. The percentage point improvement of selected and deselected solutions that are found with UDQD against baseline runs with the weight set to zero is measured. Fig.~\ref{fig:parameterization} shows those runs for the planner and control tasks, with six replicates per penalty weight setting. 

The penalty derived from the UDHM behaves in a similar fashion for both tasks, although the optimal weight is higher in the control task. Because the fitness function range is the same for both tasks, the difference has to be fully explained by the structure of the hypersurface. As was already visible in Fig.~\ref{fig:mapelites2}, the solutions are not as well separated in the case of the control task. The linear penalty function therefore has to be set to a more conservative, higher value, in order for it to be effective. The optimal weight setting for this domain is equal to 10 for the planning task and 200 for the control task.

%\FloatBarrier
\subsection{User Selection Drift in Seeding and UDHM}

In this Section, UDHM is compared against the QD seeding approach that was introduced in~\cite{Hagg2018}. It is expected to have less user selection drift, as it constrains QD to find solutions closer to those that were originally deselected by the user. The amount of user selection drift should be correlated to the mutation distance used in QD. Higher mutation distances should lead to more drift in the unconstrained seeding approach as they will allow solutions to be mutated to such a degree that they can jump over to the next basin of attraction. Because solutions are not removed from the archive when only using an UDHM adjusted objective function, the random sampling in MAP-Elites will lead to discovering new valid solutions less often than when seeding, as deselected solutions will still be picked for mutation as often as selected solutions.

The user selection drift of the seeding and UDHM approaches and their combination is evaluated by varying the QD mutation distance and evaluating the user selection drift that was defined in Section \ref{sec:decisionmetric}. The higher the mutation distance, the larger the drift is expected to be, especially when not using the UDHM to constrain QD. Neither UDHM nor seeding can fully prevent discovering novel solutions that do not fulfill the selection criterion, so drift should occur in both approaches, but drift should be lower for the UDHM model.

\begin{figure}[h!tbp]
	\centering
	\includegraphics{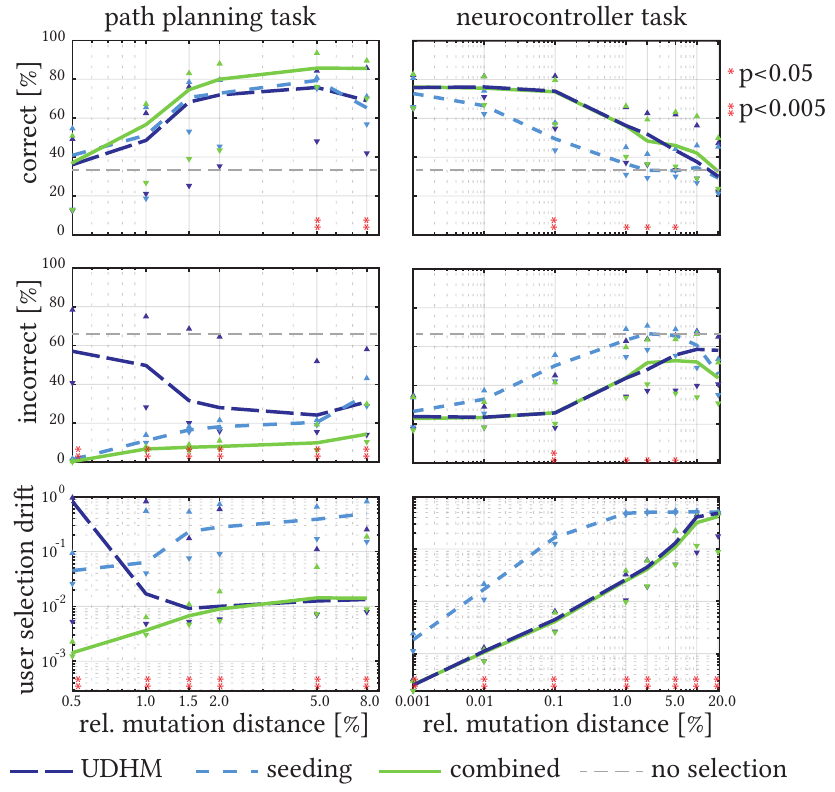}
	\caption{Median percentage of correct and incorrect solutions and median user selection drift in both tasks using UDHM, the seeding method and a combination of the two. The baseline (dotted gray lines) is uses no selection. 25\%/75\% percentile are shown as triangles. Significance results from a two-sample Kolmogorov-Smirnov test are shown with asterisks.}
	\label{fig:results}
\end{figure}

\subsubsection{Path Planning Task}

As becomes clear from the results of the planning task on the left of Fig.~\ref{fig:results}, UDHM is able to suppress user selection drift. Since UDHM starts its search at all locations in $\mathcal{S}$ and $\overline{\mathcal{S}}$, the higher the mutation distance is, the more often a deselected solution can mutate towards $\mathcal{S}$.  A large selection drift can be seen for small mutation distances when only using the UDHM, because without seeding, and with only a limited number of generations, UDQD can not mutate solutions far enough away from the deselected set. In the seeding approach QD always starts exactly at the selection $\mathcal{S}$. The approach shows less drift for very small mutation distances because it simply does not get the opportunity to mutate away far enough to generate solutions that use a different exit. It however does not find many new solutions, as it on average ends up at about the same number of correct solutions compared to not using selection at all (33\%), which is the expected value in a non-biased maze with three exits in the inner ring. With increasing mutation distance, the seeding approach drifts away from the user selection. The combination of the two approaches performs best, as the search starts at the selected locations and is suppressed from moving too far away. 

\begin{table}[htbp]
	\centering
	\begin{tabular}{ccccc}
		& none & UDHM & Seeding & Combined \\
		\noalign{\smallskip}
		\hline
		\textbf{Path planning}\\
		correct \% & 33 & 58 & 65 & \textbf{72} \\		
		incorrect \% & 66 & 39 & 18 & \textbf{7}\\			
		drift $d_{\mathcal{M}}$ & - & 0.01 & 0.17 & \textbf{0.00} \\		
		\textbf{Control}\\
		correct \% & 33 & \textbf{60} & 44 & \textbf{60} \\		
		incorrect \% & 66 & 39 & 52 & \textbf{36}\\			
		drift $d_{\mathcal{M}}$ & - & 0.03 & 0.48 & \textbf{0.02}  
	\end{tabular}
	\caption{Median correct, incorrect exits taken, and drift. Baseline results without selection under "none".}
	\label{table:results}	
\end{table}	

The median values are shown in Table~\ref{table:results}. 72\% of the solutions found take the correct exit, while only 7\% of the solutions take the wrong exit. A two-sample Kolmogorov-Smirnov test was performed to show the mutation distances at which the combined approach produces significantly different results than the seeding approach.

\subsubsection{Control Task}

The results for the non-linear and reactive control task show a qualitatively different behavior. In this case small mutation distances are beneficial to all three user selection variants. It is not hard to see why this can happen. If the weights to the output neuron that is controlling rotational movement are a bit higher, the robot will rotate more in the beginning of trajectory and might select another exit. A small change in a neural controller might lead to the robot selecting a very different path. With increasing mutation distance, the controllers move further away from the initial archive. The UDHM is able to hold the selection much longer (up to a mutation distance of 0.1\%) but at some point it gives away to the pressure exerted by mutation. The combined approach benefits only from the UDHM as it shows the same behavior.

The mean values are shown in Table~\ref{table:results}. The low mutation distances in this case still allow jumping the gap between basins of attraction because of the non-linear mapping of genotype to phenotype and behavior. This explains why all approaches perform worse than in the planning task. The combined approach performs best but very similar to the UDHM alone.

\subsection{Discussion}
The objective in the experiments does not contain any information on what exit in the inner ring should be taken. Instead, the user can select the preferred solutions. A combination of seeding and the UDHM leads to a robust selection model within UDQD. By capturing the user's selection in a model, as opposed to the seeding method that was proposed in~\cite{Hagg2018}, the UDHM adds continuous control over the QD search and less user selection drift takes place. The genotype to phenotype mapping seems to influence whether a small or large mutation distance should be used in QD. In general, the mutation distance for an unaligned genotype-to-phenotype mapping has to be lower, as similar genotypes are more likely to produce dissimilar phenotypes. Taking influence on QD search, constraining it by adjusting the objective function allows QD to find new solutions that adhere to a user's selection.

\section{Conclusions}

User selection drift is detrimental to the expected behavior of an interactive optimization algorithm. In this work UDHMs are introduced that allow modeling the state of a QD algorithm at the time the user selects from the QD archive. The user selection drift, a comparison of the distance of a candidate solution to the set of selected and to the set of deselected genotypes, is used to penalize solutions that are too close to solutions that were explicitly not selected. A user driven QD algorithm is formalized that uses the penalty in its objective function. The UDHM is compared against an approach that seeds the QD archive with the selected solutions. 

Evaluation is done in a new multimodal benchmark domain that allows comparison for two different representations in a planning and control task. The tasks allow the comparison of effects of using differently coupled representations that are hard to quantify in a pure design task, but results can be transferred back to a design case, as was shown in~\cite{Hagg2018}. Both tasks show that QD can be influenced towards a user's selection. The structure of the decision hyperspace created using t-SNE is shown to be permissive to compare solutions of various dimensionality. UDHM is most effective when combined with seeding. Depending on the representation and the genotype to behavior/phenotype mapping, the mutation distances can be high or should be more conservative. UDHM, especially when combined with the seeding approach, is able to influence QD to discover new solutions that adhere to the user's decision. Of course the occasional misclassification by the user will cause unexpected behavior, and multiple selection rounds might have to be applied. The introduced models open up the possibility of using QD in an interactive optimization process; QD can be used within the design by shopping paradigm~\cite{Balling1999}.

The limits of the approach for highly nonlinear and reactive problems might introduce the necessity to measure similarity in a different way, not based on the genotype but rather on the phenotype or behavior itself. Importantly, an interactive design process consists of more than one decision or selection. The models and the drift penalty that is introduced allow a concatenation of decision models, yet the limits of the efficacy for multiple decisions have not been evaluated. The retraction of user decisions should be possible by removing the UDHM instance from the objective function, but the effect of such an act on QD search is unknown. Further research should also be done to consider the effect of user constraints on the divergent behavior of QD. Finally, a multiobjective approach could disentangle the penalty from the fitness function, and removing the penalty weight that has to be parameterized for each domain.

Interactive control of divergent evolutionary optimization can provide intuition to engineers and designers while keeping the human in control. This makes QD more useful for real world design and engineering tasks, where understanding and solving a problem are coupled. Multimodal optimization algorithms like QD allow designers and engineers to not only discover what is possible and optimal, but to explore their preferences in an interactive setting, decomposing the twin goals of optimization and understanding.

\begin{acks}
	This work received funding from the German Federal Ministry of Education and Research, and the Ministry for Culture and Science of the state of Northrhine-Westfalia (research grants 03FH012PX5 and 13FH156IN6). 
\end{acks}

\bibliographystyle{ACM-Reference-Format}
\bibliography{geccohagg} 

\end{document}